\pdfoutput=1

\documentclass[11pt]{article}

\usepackage[final]{acl}
\usepackage{amsmath}
\usepackage{booktabs}
\usepackage{multirow}
\usepackage{graphicx}
\usepackage{colortbl}
\usepackage{natbib}
\usepackage{amssymb}

\usepackage{times}
\usepackage{latexsym}

\usepackage[T1]{fontenc}

\usepackage[utf8]{inputenc}

\usepackage{microtype}

\usepackage{inconsolata}

\usepackage{graphicx}

\title{RAVEN: Multitask Retrieval Augmented Vision-Language Learning}

\author{
 \textbf{Varun Nagaraj Rao\textsuperscript{1}\thanks{Work conducted during an internship at Amazon.}},
 \textbf{Siddharth Choudhary\textsuperscript{2}},
 \textbf{Aditya Deshpande\textsuperscript{3}\thanks{Work done while at Amazon.}},
 \textbf{Ravi Kumar Satzoda\textsuperscript{2}},
\\
 \textbf{Srikar Appalaraju\textsuperscript{2}\thanks{Correspondence to: srikara@amazon.com}}
\\
\\
 \textsuperscript{1}Princeton University,
 \textsuperscript{2}AWS AI Labs,
 \textsuperscript{3}Apple
}

\begin{document}
\maketitle
\begin{abstract}
The scaling of large language models to encode all the world's knowledge in model parameters is unsustainable and has exacerbated resource barriers.  
Retrieval-Augmented Generation (RAG) presents a potential solution, yet its application to vision-language models (VLMs) is underexplored. 
Existing methods focus on models designed for single tasks. Furthermore, they're limited by the need for resource intensive pretraining, additional parameter requirements, unaddressed modality prioritization and lack of clear benefit over non-retrieval baselines. 
This paper introduces RAVEN, a multitask retrieval augmented VLM framework that enhances base VLMs through efficient, task-specific fine-tuning. By integrating retrieval augmented samples without the need for additional retrieval-specific parameters, we show that the model acquires retrieval properties that are effective across multiple tasks.
Our results and extensive ablations across retrieved modalities for the image captioning and VQA tasks indicate significant performance improvements compared to non retrieved baselines -- +1 CIDEr on MSCOCO, +4 CIDEr on NoCaps, and nearly a +3\% accuracy on specific VQA question types. This underscores the efficacy of applying RAG approaches to VLMs, marking a stride toward more efficient and accessible multimodal learning.
\end{abstract}    
\section{Introduction}
\label{sec:intro}





The rapid growth in model sizes in NLP, as highlighted by OpenAI's LLM progression from GPT-2's 1.5 billion parameters \cite{radford2019language} to GPT-3's 175 billion \cite{brown2020language}, and further to over a trillion in GPT-4 \cite{openai2023gpt}, is a source of increasing concern. This trend requires more data and computational power, leading to higher carbon emissions and presenting significant obstacles for less-resourced researchers \cite{strubell-etal-2019-energy}. In response, the field is pivoting to approaches like Retrieval-Augmented Generation (RAG) \cite{lewis2020retrieval}, which incorporates external non-parametric world knowledge into a pretrained language model, removing the necessity of encoding all information directly into the model's parameters. However, this strategy is not yet widely applied in vision-language models (VLMs) \cite{li2022blip, wang2021simvlm, alayrac2022flamingo, chen2022pali, radford2021learning, wang2022ofa}, which process both image and textual data, and are typically more resource-intensive. Moreover, VLMs often rely on massive datasets like LAION-5B \cite{schuhmann2022laion}, presenting a significant opportunity for performance gains through retrieval augmentation.

The scant prior work exploring retrieval augmentation applied to VLMs, although promising, is beset with several limitations. Most importantly, they rely on pretraining with retrieval specific parameters \cite{hu2023reveal, ramos2023smallcap, yang2023re}; as a result the performance improvement over non-retrieval baselines cannot be established and the benefit due to retrieval augmentation cannot be independently discerned. Next, model architectures are suited to only a single task, and therefore, experimental evaluation is also only presented on a single task e.g. on image captioning \cite{ramos2023smallcap, ramos2023retrieval, yasunaga2023retrieval}; other image-to-text tasks like VQA are ignored. Further, the decision on which modality to prioritize during retrieval - textual, visual, or a combination of both - is not established. Some works \cite{yasunaga2023retrieval,chen2022murag} retrieve and concatenate both image and text, while others \cite{ramos2023retrieval, ramos2023smallcap, yang2023re} only retrieve text, even though they all evaluate on image-to-text tasks. Finally, we also observe that overlaps between the retrieval and pre-training/fine-tuning datasets exist; for example,  \citet{ramos2023retrieval, ramos2023smallcap} pretrain and retrieve from MSCOCO. This can confound the benefits attributed to the RAG approach, underscoring the need for a larger and non-overlapping external memory.


\begin{figure*}[!htb]
    \centering
    \includegraphics[width=0.93\textwidth]{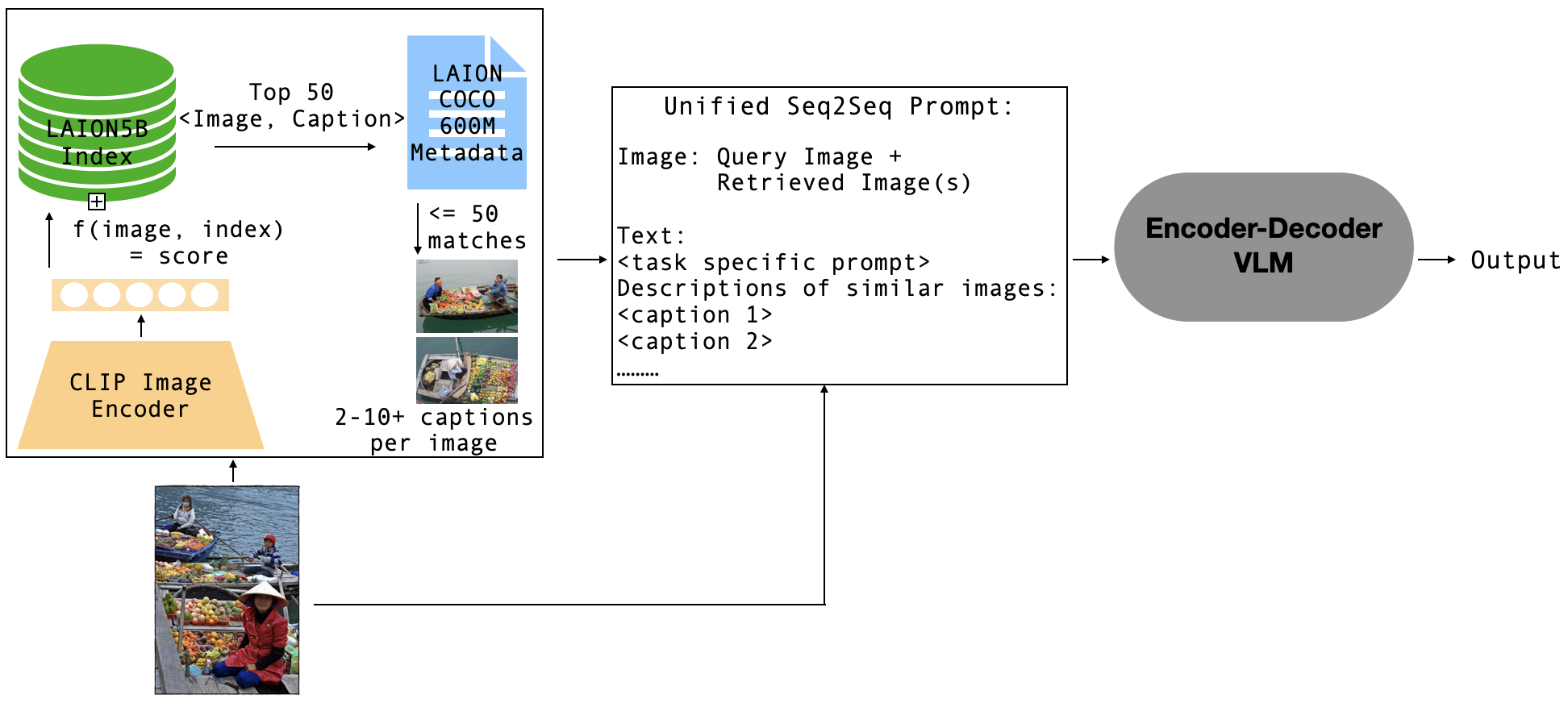}
    \caption{Illustration of our RAVEN framework. Given an input image, we retrieve image-text pairs from an external memory. Subsequently, we use a multitask pretrained base vision-language model (VLM) to encode the retrieved samples along with the query and decode to generate an output by attending over both the query and retrieved samples.}
    \label{fig:model}
\end{figure*}

In this paper, we present RAVEN (see Figure \ref{fig:model}), a multitask retrieval augmented framework adaptable to any multitask base VLM. The framework does not rely on pretraining with retrieval specific parameters, and is suitable to a variety of tasks. Importantly, the design of RAVEN allows for a comprehensive investigation of the performance benefits over non-retrieval baselines, and implications of retrieving and using different modalities. Specifically, our key contributions are as follows: 
\begin{enumerate}
    \item We are the first to design a multitask retrieval augmented VLM framework (RAVEN), which relies on only fine-tuning, no retrieval specific trainable parameters and is adaptable to any multitask base VLM.
    \item Our method allows for comprehensive ablations which examine the trade-offs between retrieval modalities and their advantages relative to non-retrieval baselines while using a non-overlapping and larger external memory. 
    \item We demonstrate the benefits and limitations of our approach on Image Captioning and VQA through quantitative and qualitative analysis. Our results achieve a new state-of-the-art performance improvement compared to non retrieved baselines: +1 CIDEr on MSCOCO, +4 CIDEr on NoCaps (using magnitudes of fewer parameters than prior works), and nearly a +3\% accuracy on specific VQA question types.
\end{enumerate}


Broadly, our work expands the empirical knowledge on RAG techniques and contributes to the rapidly growing body of work focusing on their applications to multitask VLMs. Ultimately, this work establishes a clearer understanding of the role of retrieval augmentation in VLMs, paving the way for more efficient and sustainable approaches in the field.
\section{Related Work}
\label{sec:related_work}
\subsection{Vision Language Models}
Vision language models are an emerging type of multi-modal AI system that can process both visual and textual data \cite{Appalaraju2023DocFormerv2LFAAAI, Appalaraju_2021_ICCV}
They build upon recent advances in computer vision and natural language processing to generate textual descriptions of images, answer visual questions, and perform other vision-and-language tasks. Earlier works in this direction unified multiple tasks like image captioning, image classification etc. using a simple sequence-to-sequence framework. Some notable examples include OFA~\cite{wang2022ofa}, GIT~\cite{wang2022git}, SimVLM~\cite{wang2021simvlm}. Recent vision-language models \cite{biten2022latr} augment pre-trained large language models with visual encoder. For example, Frozen~\cite{tsimpoukelli2021frozen}, Flamingo~\cite{alayrac2022flamingo}, BLIP~\cite{li2022blip}, InstructBLIP~\cite{instructblip}, LLaVA~\cite{liu2023llava}, MiniGPT-4~\cite{zhu2023minigpt}, Kosmos-1~\cite{kosmos-1}, Pali~\cite{chen2022pali}. In this work, we use OFA~\cite{wang2022ofa}  as the baseline rather than using VLMs augmented with pretrained LLMs. This choice allows us to remove the effects of in-context learning abilities of the pretrained language models from the resulting enhancement brought by retrieval-augmented vision-language modeling.

\subsection{Retrieval Augmented Generation in NLP}
Retrieval augmentation has become an important technique for improving natural language processing models. One of the first works in this area was kNN-LM by Khandelwal et al.~\cite{khandelwal20generalization} who showed how interpolating over nearest neighbors from any text collection could improve generalization. This was followed by RETRO~\cite{Borgeaud2021retro}, which scaled up the retrieval corpus to trillions of tokens. Another line of work has focused on integrating Wikipedia passages directly into models like REALM~\cite{guu2020realm}, RAG~\cite{lewis2020retrieval}, and FiD~\cite{izacard-grave-2021-leveraging}. By retrieving and conditioning on relevant Wikipedia passages, these models can better perform knowledge-intensive downstream tasks like question answering. Overall, retrieval augmentation has proven to be a highly effective way of injecting knowledge into language models to improve their capabilities. The techniques have progressed from simple corpus retrieval to integrated and scalable architectures that retrieve from large knowledge bases like Wikipedia.

\subsection{Retrieval Augmented Generation in VLMs}
Recent years have seen significant progress in extending retrieval-augmented generation to vision-language models. One of the earliest works is Multimodal Retrieval-Augmented Transformer (MuRAG) which utilizes non-parametric multimodal memory for language generation improvement~\cite{chen2022murag}.
In image-to-text generation,
Smallcap~\cite{ramos2023smallcap}, exhibits competitive performance on COCO and other domains through retrieval from target-domain data. Sarto et al.\cite{sarto2022retrieval} use kNN memory for image captioning, enhancing knowledge retrieval from external corpora.
Re-ViLM~\cite{yang2023re}, built upon the Flamingo~\cite{alayrac2022flamingo}, and supports retrieving the relevant knowledge from the external database for zero and in-context few-shot image-to-text generations. 
Recently, Iscen et al.~\cite{Iscen2023ImprovingIR} proposed to equip contrastive vision-text models with the ability to refine their embedding with cross-modal retrieved information from a memory at inference time, which greatly improved their zero-shot predictions. 
Hu et al.~\cite{hu2023reveal} presented REVEAL that learns to encode world knowledge into a large-scale memory, and to retrieve from it to answer knowledge-intensive queries, and achieves state-of-the-art results on visual question answering and image captioning. 
In text-to-image generation, Chen et al.~\cite{Chen2022ReImagenRT} presented Re-Imagen that uses retrieved information to produce high-fidelity and faithful images, even for rare or unseen entities. RA-CM3 is the first multimodal model that can retrieve and generate mixtures of text and images and exhibits novel capabilities such as knowledge-intensive image generation and multi-modal in-context learning~\cite{yasunaga2023retrieval}. 

Our multitask framework, RAVEN, extends beyond RA-CM3 by supporting both captioning and VQA, and it diverges from REVEAL \cite{hu2023reveal} by attaining retrieval capabilities solely through fine-tuning, eliminating the need for pretraining and additional retrieval-specific parameters; and is adaptable to any base VLM.

\section{Proposed Approach}
\label{sec:approach}
\subsection{RAVEN Framework}
Our framework, RAVEN, is illustrated in Figure \ref{fig:model}. At a high level, given a multimodal input consisting of images and text, we use a \textbf{retriever} to retrieve relevant image-text pairs from a large external memory. Subsequently, we use a pretrained multitask encoder-decoder \textbf{VLM} which refers to the retrieved context in addition to the multimodal query and generates a textual output. Importantly, we demonstrate that through short, but efficient, task specific fine-tuning of the base VLM, with concatenated retrieval augmented samples and no additional retrieval-specific trainable parameters, the model acquires retrieval properties which generalizes to multiple tasks. We now describe both these components in detail. 
\subsection{Multimodal Retriever}
Our semantic search based retrieval system, relies on the Facebook AI Similarity Search (FAISS) library \cite{douze2017faiss}. FAISS enables high-dimensional vector indexing within an external memory and facilitates efficient search through an approximate nearest neighbor approach based on a specified similarity measure, such as dot-product similarity. We utilize the publicly available Laion-5B \cite{schuhmann2022laion} image-based index which consists of 5 billion images and corresponding alt text. 

To describe the retrieval steps in detail, we first encode the query image using a CLIP-based image encoder \cite{radford2021learning} into a dense vector. Next, we follow the Dense Retrieval method outlined in \citet{karpukhin2020dense} to retrieve the top `k' (k can be specified by the user) image-text pairs by scoring the query (image) and memory data as follows:
\begin{equation}
    \text{score(query, memory)} = E(\text{query})^T E(\text{memory})
\end{equation}
where E is the CLIP-based image encoder. Finally, we perform Maximum Inner Product Search (MIPS) over the memory to obtain the top `k' candidate image-text pairs sorted according to the score. 

Our retrieval approach ensures that the retrieved samples, which are provided as additional context to the model, along with the query image, are \textit{relevant}, \textit{diverse} and in the \textit{style} of our target datasets.  Relevance is easily ensured through sampling based on the top similarity score. However, simply sampling based on relevance score can result in exact or near duplicates resulting in poor performance. To avoid this redundancy and enhance diversity, we exclude near duplicate images. Finally, to use COCO-style captions rather than the noisy image alt text in Laion-5B, we map the retrieved samples from Laion-5B down to the Laion-COCO 600M \footnote{\url{https://laion.ai/blog/laion-coco/}} subset, whose captions are synthetically generated using a BLIP model trained on COCO-style captions. This can result in some missing data due to lack of matches with LAION-COCO 600M and also due to failure of LAION-COCO 600M raw image downloads. Our approach is robust to these missing samples.   

\subsection{Base Vision-Language Model (VLM)}
RAVEN relies on a multitask, multimodal encoder-decoder base VLM which can easily leverage additional multimodal context from an external memory. 

\noindent \textbf{Architecture.} \qquad For image encoding, we use a ResNet, and for text encoding we use a byte-pair encoding (BPE) to convert the text sequence into a subword sequences, and then embed them into features. We adopt a unified vocabulary encompassing linguistic and visual tokens, incorporating subwords, image codes, and location tokens.
The base architecture is the transformer; this serves as the backbone for the encoder-decoder framework. 
To enhance stability and hasten convergence, the model uses head scaling for self-attention, post-attention layer normalization (LN), and LN following the first layer of FFN. For positional information, separate absolute position embeddings are used for text and images. Notably, we decouple position correlation from token embeddings and patch embeddings, while employing 1D relative position bias for text and 2D relative position bias for images.

\noindent \textbf{VL Tasks.} \qquad All cross-modal tasks are cast as Seq2Seq generation. We focus on 2 popular image-to-text tasks, image captioning and visual question answering (VQA). For image captioning, the model adeptly adopts the Seq2Seq format, generating captions based on both the provided image and the input textual prompt, ``What does the image describe?''. For VQA, the model takes in the image and the question as inputs, learning to generate accurate responses.

\noindent \textbf{Need for Retrieval in VL tasks.} \qquad Retrieval can benefit performance in VL tasks as contextual information can be crucial for guiding models to accurate answers. Moreover, the retrieval mechanism can mitigate bias by sourcing information from diverse datasets, countering the influence of biased training data. Specifically, in VQA, image content, such as object attributes, strongly correlates with questions and answers, making captions valuable auxiliary information while similar/retrieved images are less informative \cite{gur2021cross}. In captioning, additional textual context resembles few-shot inference \cite{yasunaga2023retrieval}.

\noindent \textbf{Reasons for OFA\cite{wang2022ofa} as a VLM backbone.} \qquad We list 4 reasons for choosing OFA rather than alternates like Beit-3 \cite{beit3} and Open Flamingo \cite{awadalla2023openflamingo}: \textit{First}, OFA is naturally suited to our approach as it unifies multiple modalities and tasks into a single Seq2Seq model; the multitask backbone is a deliberate design choice that underscores the versatility of our approach and is a foundational element crucial to our model's architecture. \textit{Second}, we can easily endow the model retrieval augmented capabilities through short, but efficient, task specific fine-tuning with no additional trainable parameters. Moreover, we intentionally avoided recent MLLM models like LLaVa or Flamingo which contain an LM to not add additional trainable parameters, remove their in-context learning ability and isolate retrieval capabilities within an encoder-decoder backbone, a first in the field.
\textit{Third}, the codebase is open source, modular and easy to extend. \textit{Finally}, the base OFA model is not very large (182M parameters) given our compute and finance limitations, but sufficient to demonstrate the benefits of our framework. 
\section{Experiments}
\label{sec:expts}
In this section, we evaluate the performance of our approach under the fine-tuning setting on various image captioning and VQA benchmarks. We aim to demonstrate the benefits of retrieval augmentation on the generated captions and answers through retrieving relevant knowledge from a large external non-overlapping database with the fine-tuning datasets. Our experiments show clear benefits of our approach compared to non-retrieval baselines. Furthermore, the performance is competitive with similarly sized models, and even exceeds the performance of existing widely used captioning and VQA models several magnitudes larger.

\subsection{Training Setup}

\subsubsection{Data} 
We make use of an external memory and task specific fine-tuning datasets in our implementation. For captioning, we use the MSCOCO 2014 Karpathy Splits for fine-tuning and NoCaps for a zero-shot evaluation. For VQA, we use the VQA v2 dataset augmented with VG-QA questions during fine-tuning. We use Laion-5B index as our external memory and map down to Laion-COCO 600M subset to retrieve image-caption pairs. The datasets are summarized in Table \ref{tab:caption_data_summary} and \ref{tab:vqa_data_summary}. Notably, unlike prior work, we ensure the fine-tuning datasets and external memory do not have any overlap, to realize the true benefits of retrieval augmentation in practical settings.
\begin{table}[h]
\resizebox{\columnwidth}{!}{%
\begin{tabular}{@{}c|c|cccc@{}}
\toprule
\textbf{Dataset} &
  \textbf{Split} &
  \textbf{\begin{tabular}[c]{@{}c@{}}\# of images\\ (original)\end{tabular}} &
  \textbf{\begin{tabular}[c]{@{}c@{}}\# of images\\ (caption)\end{tabular}} &
  \textbf{\begin{tabular}[c]{@{}c@{}}\# of images\\ (caption + image)\end{tabular}} &
  \textbf{\begin{tabular}[c]{@{}c@{}}Size - w or w/o \\ retrieval\end{tabular}} \\ \midrule
\multicolumn{1}{c|}{\multirow{3}{*}{\begin{tabular}[c]{@{}c@{}}MSCOCO\\ Karpathy \\ Split (2014)\end{tabular}}} &
  \multicolumn{1}{c|}{train} &
  113287 &
  108780 &
  107800 &
  37G / 64G \\
\multicolumn{1}{c|}{}       & \multicolumn{1}{c|}{val}  & 5000 & 4776 & 4725 & 330M / 573M \\
\multicolumn{1}{c|}{}       & \multicolumn{1}{c|}{test} & 5000 & 4817 & 4778 & 329M / 576M \\ \midrule
\multicolumn{1}{c|}{NoCaps} & \multicolumn{1}{c|}{val}  & 4500 & 4275 & 4239 & 295M / 512M \\ \bottomrule
\end{tabular}%
}
\caption{Captioning dataset summary}
\label{tab:caption_data_summary}
\end{table}
\begin{table}[htb]
\resizebox{\columnwidth}{!}{%
\begin{tabular}{@{}c|ccccc@{}}
\toprule
\textbf{Split} &
  \textbf{\begin{tabular}[c]{@{}c@{}}\# of samples\end{tabular}} &
  \textbf{\begin{tabular}[c]{@{}c@{}}\# of images\\ (original)\end{tabular}} &
  \textbf{\begin{tabular}[c]{@{}c@{}}\# of images\\ (caption)\end{tabular}} &
  \textbf{\begin{tabular}[c]{@{}c@{}}\# of images\\ (caption + image)\end{tabular}} &
  \textbf{\begin{tabular}[c]{@{}c@{}}Size - w or w/o \\  retrieval\end{tabular}} \\ \midrule
train & 1,358,769  & 121,277 & 116,439 & 115,387 & 106G / 151G \\
val  & 10,402   & 2,000   & 1,924   & 1,906   & 653M / 1.2G \\
test-dev & 107,394 & 36,807  & 35,107  & 34,760  & 28G / 50G   \\
test-std & 447,793 & 81,434  & 77,856  & 77,098  & 28G / 50G   \\ \bottomrule
\end{tabular}%
}
\caption{VQA v2 dataset summary}
\label{tab:vqa_data_summary}
\end{table}

\paragraph{Missing Samples:} Retrieved data can be missing for 2 reasons: (1) lack of matches of the Laion-5B retrieved samples with the Laion-COCO 600M subset, and (2) raw image download failure. For captioning, we only work on the subset of samples which have both retrieved captions and images. We validate that augmentation with images is not useful, and subsequently decide to only use retrieved captions for augmentation. For VQA, we retain the original dataset, and missing captions are handled with an empty string. This allows us to evaluate our results on the VQA evaluation server. Importantly, the model learns to be robust to samples which may not have corresponding retrieved context at inference; a scenario common in practice.

\subsubsection{Implementation}
Our retriever uses the off-the-shelf CLIP image encoder \cite{radford2021learning} for both the query and memory encoders. We use FAISS \cite{douze2017faiss} to index the external Laion-5B image-based memory and perform MIPS-based top-50 retrieval. We then map down to the Laion-COCO 600M subset ensuring to select, when it exists, the top-1 image (excluding exact or near duplicates), and all associated metadata, including the top caption, all captions and alt text. The retrieved samples are concatenated with the original samples in the TSV file provided as input during the fine-tuning process.

We ensure our fine-tuning process is able to operate in resource constrained settings. We use a lightweight OFA-base \cite{wang2022ofa} model checkpoint of 182M parameters as our multitask VLM. The maximum sequence length is 1024. We fine-tune the model for 8-12 hours, upto 10 epochs, on 4 V100 32GB GPU's. Our implementation is in PyTorch. We increase the max source length from 80 upto 600 to account for the retrieved samples. Otherwise, we rely on the task-specific default hyperparameters in the OFA-base run scripts. 

Following the OFA implementation, we optimize the model with the standard cross-entropy loss. Given an input image i, a prompt t, and an output y, we minimize the loss $L = - \sum_{j=i}^{|y|} \log P_{\theta}(y_j|y<j,i,t)$ where $\theta$ refers to the model parameters. For inference, we decode using beam search, to enhance the quality of generation. For the VQA task, we employ a trie-based search to only search over a bounded set of vocabulary (top 3129 VQA v2 answers) to prevent labels out of the closed label set during inference.

\subsection{Evaluation Setup}

\subsubsection{Baselines}

We establish baselines to gauge the performance of RAVEN in comparison to various configurations:

\noindent \textbf{Captioning.} \quad (1) Retrieval Only: This baseline involves using the top caption retrieved from the memory as the generated output. It serves as a benchmark to assess the additional benefits gained through fine-tuning the OFA-base model. (2) Zero Shot In-Context Retrieval: During inference, this baseline directly concatenates the retrieved top caption and all captions with the prompt. The objective is to evaluate the model's capacity to leverage retrieved context without any pretraining or fine-tuning. (3) No Retrieved Samples: In this scenario, the model undergoes fine-tuning solely on the target dataset without incorporating any retrieved context. This baseline helps establish a performance reference point.

\noindent \textbf{VQA.} \quad No Retrieved Samples: Similar to the captioning task, this baseline involves fine-tuning the model exclusively on the target dataset without incorporating any retrieved context.

In all cases, we report performance gains relative to the ``No Retrieved Samples'' baselines to highlight the efficacy of our proposed approach. Notably, most prior work fail to report this baseline making it challenging to assess the benefits of retrieval augmentation. 

Additionally, we provide a comparative analysis by reporting recent baselines and the current State-of-the-Art (SOTA) for both captioning and VQA tasks. This comparative assessment considers performance metrics and the number of parameters, offering a comprehensive view of the landscape and positioning our model within the current state-of-the-art research.
\subsubsection{Metrics}
In evaluating the performance of RAVEN for captioning, we employ two key metrics: BLEU@4 and CIDEr. BLEU@4 measures the quality of generated captions by assessing the overlap of n-grams (in this case, four-grams) between the generated caption and reference captions. Meanwhile, the CIDEr metric gauges the diversity and distinctiveness of generated captions by considering consensus across multiple reference captions.

For the VQA task, we utilize accuracy as the evaluation metric. This measure is computed using the \texttt{Eval.ai} server.

\subsubsection{Ablations}
\begin{table*}[!htb]
\resizebox{\textwidth}{!}{%
\begin{tabular}{@{}ccclccc@{}}
\toprule
\multicolumn{2}{c|}{\textbf{\begin{tabular}[c]{@{}c@{}}Retrieval\\ Modality\end{tabular}}} &
  \multicolumn{1}{c|}{} &
  \multicolumn{1}{c|}{} &
  \multicolumn{2}{c|}{\textbf{MSCOCO}} &
  \textbf{NoCaps} \\ \cmidrule(r){1-2} \cmidrule(l){5-7} 
\textbf{Image} &
  \multicolumn{1}{c|}{\textbf{Text}} &
  \multicolumn{1}{c|}{\multirow{-2}{*}{\textbf{\begin{tabular}[c]{@{}c@{}}\# of\\ Parameters\end{tabular}}}} &
  \multicolumn{1}{c|}{\multirow{-2}{*}{\textbf{Ablation Description}}} &
  \textbf{BLEU@4} &
  \multicolumn{1}{c|}{\textbf{CIDEr}} &
  \textbf{CIDEr} \\ \midrule
  \multicolumn{7}{c}{\cellcolor[HTML]{EFEFEF}\textbf{Our Approach (Image, Text, Image+Text Retrieval)}} \\ \midrule
- &
  \multicolumn{1}{c|}{-} &
  \multicolumn{1}{c|}{-} &
  \multicolumn{1}{l|}{retrieval only} &
  0.1905 &
  \multicolumn{1}{c|}{74.98} &
  71.68 \\
- &
  \multicolumn{1}{c|}{-} &
  \multicolumn{1}{c|}{182M} &
  \multicolumn{1}{l|}{zero shot in-context retrieval with top caption + all captions} &
  0.3777 &
  \multicolumn{1}{c|}{128.91} &
  103.99 \\
- &
  \multicolumn{1}{c|}{-} &
  \multicolumn{1}{c|}{182M} &
  \multicolumn{1}{l|}{no retrieved samples} &
  0.4102 &
  \multicolumn{1}{c|}{\textbf{137.25}} &
  \textbf{106.69} \\ \midrule
- &
  \multicolumn{1}{c|}{\checkmark} &
  \multicolumn{1}{c|}{182M} &
  \multicolumn{1}{l|}{top caption} &
  0.4102 &
  \multicolumn{1}{c|}{\textbf{138.23* (+0.98)}} &
  109.76 \\
- &
  \multicolumn{1}{c|}{\checkmark} &
  \multicolumn{1}{c|}{182M} &
  \multicolumn{1}{l|}{alt text} &
  0.4125 &
  \multicolumn{1}{c|}{137.19} &
  106.81 \\
- &
  \multicolumn{1}{c|}{\checkmark} &
  \multicolumn{1}{c|}{182M} &
  \multicolumn{1}{l|}{all captions concatenated} &
  0.4057 &
  \multicolumn{1}{c|}{137.70} &
  109.72 \\
- &
  \multicolumn{1}{c|}{\checkmark} &
  \multicolumn{1}{c|}{182M} &
  \multicolumn{1}{l|}{top caption + all captions} &
  0.4108 &
  \multicolumn{1}{c|}{\textbf{138.17* (+0.92)}} &
  \textbf{111.00 (+ 4.31)} \\
- &
  \multicolumn{1}{c|}{\checkmark} &
  \multicolumn{1}{c|}{182M} &
  \multicolumn{1}{l|}{top caption + all captions + alttext} &
  0.4104 &
  \multicolumn{1}{c|}{138.03} &
  109.88 \\ \midrule
\checkmark &
  \multicolumn{1}{c|}{-} &
  \multicolumn{1}{c|}{182M} &
  \multicolumn{1}{l|}{image} &
  0.4087 &
  \multicolumn{1}{c|}{136.95} &
  106.22 \\ \midrule
\checkmark &
  \multicolumn{1}{c|}{\checkmark} &
  \multicolumn{1}{c|}{182M} &
  \multicolumn{1}{l|}{image + top caption + all captions} &
  0.4081 &
  \multicolumn{1}{c|}{136.85} &
  107.28 \\ \midrule
\multicolumn{7}{c}{\cellcolor[HTML]{EFEFEF}\textbf{Image Captioning Baselines (Fine-tuning)}} \\ \midrule
- &
  \multicolumn{1}{c|}{-} &
  \multicolumn{1}{c|}{420M} &
  \multicolumn{1}{l|}{Re-ViLM (base, \cite{yang2023re})} &
  0.378 &
  \multicolumn{1}{c|}{129.1} &
  105.2 \\
- &
  \multicolumn{1}{c|}{-} &
  \multicolumn{1}{c|}{364M} &
  \multicolumn{1}{l|}{Flamingo (base, re-implementation from \cite{yang2023re})} &
  0.370 &
  \multicolumn{1}{c|}{128.0} &
  102.8 \\
- &
  \multicolumn{1}{c|}{-} &
  \multicolumn{1}{c|}{252M} &
  \multicolumn{1}{l|}{$\text{BLIP}_\text{CapFilt-L}$ \cite{li2022blip}} &
  0.404 &
  \multicolumn{1}{c|}{136.7} &
  113.2 \\
- &
  \multicolumn{1}{c|}{-} &
  \multicolumn{1}{c|}{172M} &
  \multicolumn{1}{l|}{VL-T5 \cite{cho2021unifying}} &
  0.346 &
  \multicolumn{1}{c|}{116.1} &
  4.4 \\
- &
  \multicolumn{1}{c|}{-} &
  \multicolumn{1}{c|}{1.4B} &
  \multicolumn{1}{l|}{SimVLM (huge, \cite{wang2021simvlm})} &
  0.406 &
  \multicolumn{1}{c|}{143.3} &
  110.3 \\
- &
  \multicolumn{1}{c|}{-} &
  \multicolumn{1}{c|}{5.1B} &
  \multicolumn{1}{l|}{GIT2 (current SOTA \cite{wang2022git})} &
  0.432 &
  \multicolumn{1}{c|}{146.4} &
  126.9 \\ \bottomrule
  \multicolumn{7}{l}{\small *Gain with respect to the non retrieved baseline is comparable to the only prior work which reported it for the MSCOCO captioning task \cite{sarto2022retrieval}}
\end{tabular}}
\caption{Fine-tuning evaluation results using cross-entropy optimization on MSCOCO, and NoCaps benchmarks, compared with different image captioning baselines. For NoCaps, we finetune on MSCOCO karpathy train following prior works \cite{li2022blip}, and perform zero-shot evaluation. We use the Laion-5B image index mapped down to the Laion-COCO 600M subset as our external memory. We report BLEU@4 and CIDer scores for different methods and show the gain in the best performing models compared to the non-retrieved baseline.}
\label{tab:captioning_results}
\end{table*}
\begin{table*}[!htb]
\centering
\resizebox{\textwidth}{!}{%
\begin{tabular}{@{}clcccc@{}}
\toprule
\multicolumn{1}{c|}{}     & \multicolumn{1}{c|}{}                                 & \multicolumn{4}{c}{\textbf{Test-Dev Accuracy \%}} \\ \cmidrule(l){3-6} 
\multicolumn{1}{c|}{\multirow{-2}{*}{\textbf{\begin{tabular}[c]{@{}c@{}}\# of \\ Parameters\end{tabular}}}} &
  \multicolumn{1}{c|}{\multirow{-2}{*}{\textbf{Ablation Description}}} &
  \textbf{number} &
  \textbf{other} &
  \textbf{yes/no} &
  \textbf{overall} \\ \midrule
\multicolumn{6}{c}{\cellcolor[HTML]{EFEFEF}\textbf{Our Approach (Text Retrieval)}}                                                    \\ \midrule
\multicolumn{1}{c|}{182M}    & \multicolumn{1}{l|}{no retrieved samples}                     & \textbf{58.55}      & \textbf{67.47}      & \textbf{90.12}      & \textbf{75.89}      \\ \midrule
\multicolumn{1}{c|}{182M} & \multicolumn{1}{l|}{alttext}                          & 61.10       & 67.94      & 90.10       & 76.29      \\
\multicolumn{1}{c|}{182M} & \multicolumn{1}{l|}{alttext + all captions}           & 57.84      & 67.92      & 90.46      & 76.06      \\
\multicolumn{1}{c|}{182M} &
  \multicolumn{1}{l|}{top caption + all captions} &
  \textbf{\begin{tabular}[c]{@{}c@{}}61.33*   (+ 2.78\%)\end{tabular}} &
  \textbf{\begin{tabular}[c]{@{}c@{}}68.27*(+ 0.80\%)\end{tabular}} &
  \textbf{\begin{tabular}[c]{@{}c@{}}90.54*  (+0.42\%)\end{tabular}} &
  \textbf{\begin{tabular}[c]{@{}c@{}}76.75*  (+0.86\%)\end{tabular}} \\ \midrule
\multicolumn{6}{c}{\cellcolor[HTML]{EFEFEF}\textbf{VQA Baselines (Fine-tuning)}}                                                                    \\ \midrule

\multicolumn{1}{c|}{122M} & \multicolumn{1}{l|}{UnifiedVLP \cite{zhou2020unified}} &     52.10       &     60.30       &     87.20       & 70.50      \\
\multicolumn{1}{c|}{252M} & \multicolumn{1}{l|}{$\text{BLIP}_\text{CapFilt-L}$ \cite{li2022blip}} &    -        &    -        &      -      & 78.25      \\
\multicolumn{1}{c|}{1.4B} & \multicolumn{1}{l|}{SimVLM (huge, \cite{wang2021simvlm})} &     -       &       -     &       -     & 80.30       \\
\multicolumn{1}{c|}{80B}  & \multicolumn{1}{l|}{Flamingo \cite{alayrac2022flamingo}}  &       -     &       -     &        -    & 82.00       \\
\multicolumn{1}{c|}{55B}  & \multicolumn{1}{l|}{PaLI-X (2023) - current SOTA \cite{chen2022pali}}     &       -     &     -       &      -      & 86.10       \\ \bottomrule
\multicolumn{6}{l}{\footnotesize *Gain with respect to the non retrieved baseline surpasses that of the only prior work which reported it for the VQA v2 task \cite{gur2021cross}}
\end{tabular}}%
\caption{Finetuning evaluation results on VQA v2 benchmarks, compared with the non retrieval VQA baseline. We finetune our method on VQA v2 train split using a subset of the OFA dataset. We report Test-Dev accuracy \% from the eval.ai server for different methods.}
\label{tab:vqa_results}
\end{table*}

\begin{figure*}[!htb]
    \centering
    \includegraphics[width=\textwidth]{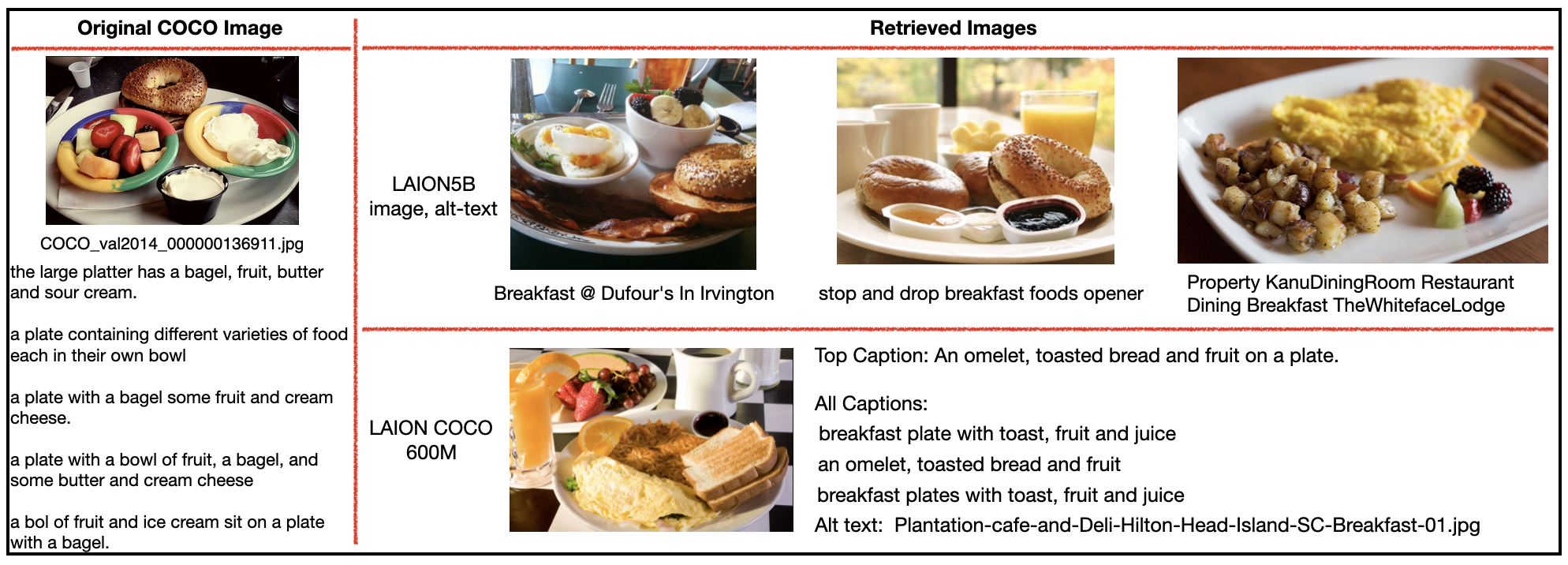}
    \caption{Examples of the retriever output given a query image.}
    \label{fig:retriever_output}
\end{figure*}

We explore three distinct sets of ablations for both captioning and VQA: text-only, image-only, and combined image and text. To the best of our knowledge, we are the first to comprehensively discern the impact of text and image modalities in retrieval augmented VLMs, providing valuable insights to model practitioners.

\noindent \textbf{Captioning.} \quad For the text-only ablation, we experiment with various combinations, concatenating one or more of the top caption, all captions, and image alt text. This helps us discern the impact of textual information in isolation. In the image-only ablation, we alter the patch size, doubling it, and employ a horizontal concatenation strategy. If a retrieved image is present, we concatenate it with the query image. In cases where the retrieved image is absent, we duplicate the query image. This analysis provides valuable insights into the model's reliance on visual information alone.
For the combined image and text ablation, we adopt a similar approach to the image-only case for processing images. Simultaneously, we concatenate the top caption and all captions to the text prompt. This exploration allows us to understand the synergistic effects of both modalities.

\noindent \textbf{VQA.}\quad Building on insights gained from the captioning task, where naive image fusion through concatenation proved less useful (see Table \ref{tab:captioning_results}), we hypothesize that captions serve as good auxiliary information in image-to-text tasks, while similar/retrieved images are less informative, since the content of the image and the objects contained is often very correlated with the question and answer. Therefore, in the VQA ablations, we exclusively consider text concatenation scenarios. This involves combining one or more of the top caption, all captions, and alt text when available. In instances where the retrieved sample is missing, we concatenate with an empty string.

\section{Results}

\subsection{Quantitative Analysis}

\noindent \textbf{Captioning.} \quad The results for image captioning, presented in Table \ref{tab:captioning_results}, reveal notable insights. Baseline comparisons indicate that both the retrieval-only and zero-shot in-context retrieval fall short of the no-retrieved samples baseline, underscoring the value of fine-tuning on the target dataset. The absence of zero-shot in-context retrieval capabilities may be attributed to the absence of a language model in the transformer-based encoder-decoder VLM architecture. In the text-only ablation, concatenating with the top caption and/or all captions yields optimal performance, demonstrating a gain of nearly 1 CIDEr point on MSCOCO and up to 4 CIDEr points on zero-shot NoCaps. The gain with respect to the non retrieved baseline is comparable to the only prior work which reported it (+1.2 CIDEr score) for the MSCOCO captioning task \cite{sarto2022retrieval}. This emphasizes the valuable contextual information provided by retrieved captions. However, concatenating with alt text proves less effective due to its inherent noise. Both image-only and combined image and text concatenation exhibit performance below the non-retrieved baseline, suggesting that retrieved images and naive concatenation introduce noise rather than relevant context. In fine-tuning settings, our model performs competitively with similar-sized models such as BLIP. Notably, in the zero-shot setting on NoCaps, our model surpasses SimVLM (1.4B vs 182M parameters), achieving a CIDEr score of 111.0 compared to 110.3.

\noindent \textbf{VQA.} \quad
Given the limited efficacy observed in the use of retrieved image for captioning (see Table \ref{tab:captioning_results}), we exclusively explore text augmentation strategies for VQA. The results, presented in Table \ref{tab:vqa_results}, align with the captioning outcomes, affirming the efficacy of text-only augmentation. Notably, across all question categories, text-only augmentation yields improvements in accuracy ranging from 0.42\% to 2.78\%. The gain with respect to the non retrieved baseline surpasses that of the only prior work which reported it (+0.36\% accuracy) for the VQA v2 task \cite{gur2021cross}. The highest performance is achieved through concatenating the top caption and all captions with the question, while the addition of alt text introduces noise, resulting in lower performance. The overall performance of our model in VQA remains competitive and comparable to similar-sized models, underscoring its robustness in leveraging textual information for accurate question answering.

\subsection{Qualitative Analysis}
In this section, we present qualitative examples that elucidate the efficacy and limitations of our approach.

\begin{figure}[!htb]
    \centering
    \includegraphics[width=\columnwidth]{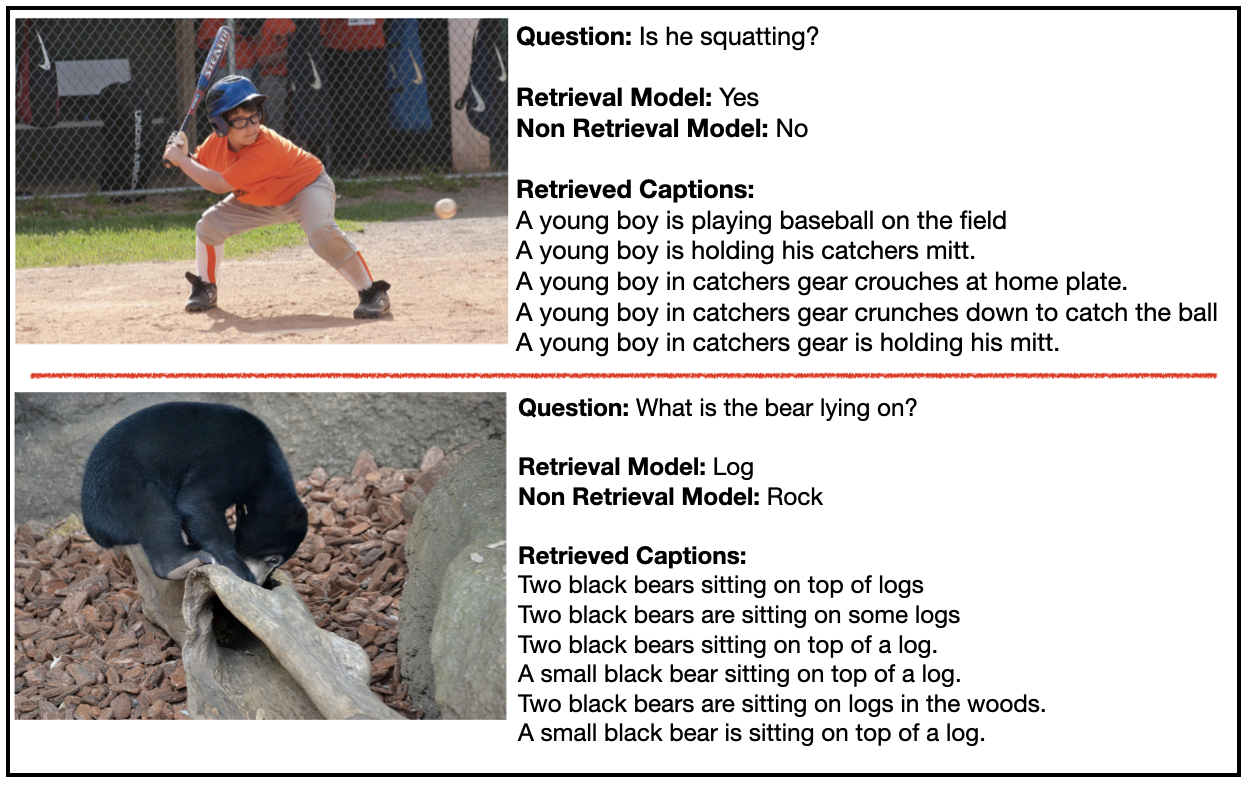}
    \caption{Examples where RAVEN succeeds in generating the correct answer.}
    \label{fig:vqa_success}
\end{figure}

\noindent \textbf{Retriever Output.} \quad Figure \ref{fig:retriever_output} illustrates the output of the retriever for a given query image. The retrieved images align with the query image, emphasizing relevance. However, Laion-5B's image alt text is observed to be noisy and differs from the required COCO-style captions. Mapping down to synthetically generated BLIP captions from the LAION-COCO 600M subset, mitigates the style issue by mimicking the COCO caption style, and offers more valuable context to the model.

\noindent \textbf{Incorporating World Knowledge.} \quad Figure \ref{fig:vqa_success} demonstrates VQA outputs leveraging world knowledge. The model adeptly utilizes entity-rich captions from the retriever to disambiguate between entities, as seen in the bear image distinguishing logs from rocks. Additionally, the model accurately identifies nuanced details, such as a boy squatting while playing baseball, by leveraging relevant context in the captions, such as the term ``crouches."

\begin{figure}[!htb]
    \centering
    \includegraphics[width=\columnwidth]{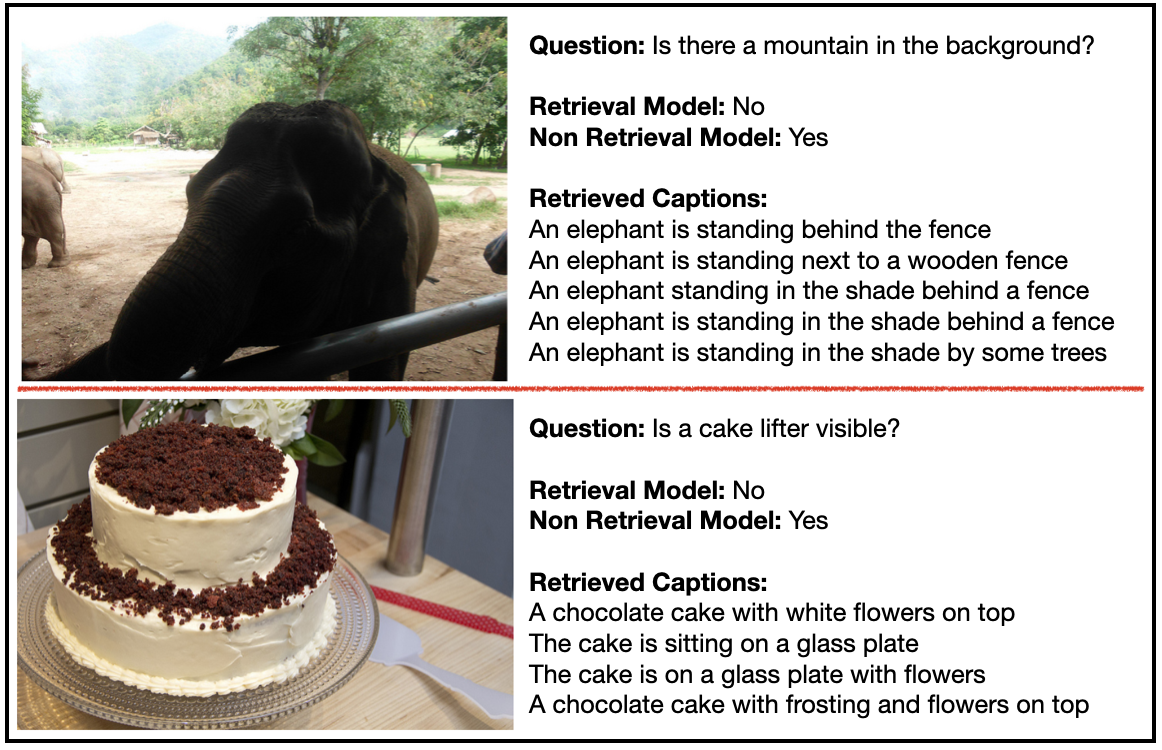}
    \caption{Examples where RAVEN fails in generating the correct answer.}
    \label{fig:vqa_failure}
\end{figure}

\noindent \textbf{Retriever Failures.} \quad Despite successes, retrieved context may not consistently contribute to specific questions, particularly when inquiries concern entities not prominently featured in the image. This issue is more pronounced in tasks such as VQA, rather than in captioning, where general knowledge about the image is often sufficient to generate high quality and diverse captions. Illustrated in Figure \ref{fig:vqa_failure}, failure cases for VQA depict relevant but insufficiently informative captions. For instance, captions for an elephant image focus on the foreground elephant, neglecting details about the background mountains and forest. Similarly, captions for a cake image lack information about the cake lifter in the corner.

\begin{figure}[!htb]
    \centering
    \includegraphics[width=0.7\columnwidth]{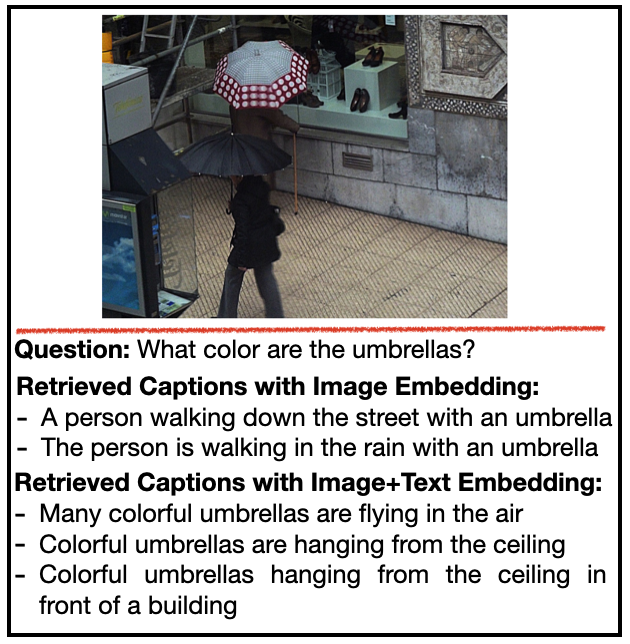}
    \caption{An example depicting the benefits of using a multimodal query embedding (average of image and question embedding). This results in the retrieval of captions relevant to both the image and question.}
    \label{fig:multimodal_query_embedding}
\end{figure}

\noindent \textbf{Multimodal Query Embedding.} \quad Considering scenarios where retrieved context may lack specificity, we propose the joint use of image and text modalities as input to the retriever, when available. Figure \ref{fig:multimodal_query_embedding} demonstrates an example where creating a multimodal query embedding by averaging image and text embeddings separately results in relevant captions addressing both the image and the question. Comprehensive exploration of scenarios where specific entity properties lack corresponding captions is deferred to future work.

\section{Conclusion and Future Work}
To address escalating model size and computational demands, we propose a retrieval augmentation framework, an alternative to storing extensive world knowledge within model parameters. Our contributions introduce a multitask, multimodal retrieval-augmented vision-language model, demonstrating adaptability across multiple tasks through computationally efficient task-specific fine-tuning. Utilizing concatenated multimodal retrieval-augmented samples from an external non-overlapping memory, without additional trainable parameters, our single model acquires robust retrieval properties. This showcases benefits in both captioning and VQA tasks using a unified approach. Notably, extensive ablations across text, image, and image-text modalities, systematically compared against non-retrieved baselines, provide valuable insights. Our findings underscore that retrieval augmentation, particularly with text in image-to-text tasks, optimally enhances performance, especially in the zero-shot setting.

Future directions involve refining sampling strategies for enhanced diversity, exploring alternative image fusion approaches, and investigating a mixture of experts to afford the model flexibility in leveraging retrieved context. Additionally, we propose extending retrieval over a composite index (image+text) to further optimize performance.

\section{Limitations}
We use a relatively small model to demonstrate performance on 2 tasks. While we acknowledge the demonstrating our approach on more tasks and larger models would be beneficial, we defer this to future work due to compute and financial constraints. RAVEN's current capability to handle diverse tasks like image captioning and VQA within a single model framework already stands as a significant advancement; and is sufficient to demonstrate the benefit of our framework.

\bibliography{references}




\end{document}